\documentclass[11pt]{article}

\usepackage{acl}

\usepackage{times}
\usepackage{latexsym}
\usepackage[T1]{fontenc}
\usepackage[utf8]{inputenc}
\usepackage{microtype}
\usepackage{inconsolata}
\usepackage{graphicx}
\usepackage{subcaption}
\usepackage{xspace}
\usepackage{CJKutf8}
\usepackage{booktabs}
\usepackage{longtable}
\usepackage{tabularx}
\usepackage{pifont}%
\usepackage{tcolorbox}        %
\tcbset{
  colback=white,              %
  colframe=black,             %
  boxrule=0.8pt,              %
  arc=2mm,                    %
  left=4pt,right=4pt,top=4pt,bottom=4pt  %
}

\def\SampleNum{2,610\xspace}
\def\TopicNum{16\xspace}
\def\LangNum{14\xspace}
\def\RegionNum{51\xspace}
\def\BenchName{CulturALL\xspace}

\title{\BenchName: Benchmarking Multilingual and Multicultural Competence of LLMs on Grounded Tasks}

\author{
    Peiqin Lin$^1$, Chenyang Lyu$^1$, Wenjiang Luo$^2$, Haotian Ye$^3$, Md Mehrab Hossain$^5$, Chunlan Ma$^3$,\\
    \textbf{Shaoxiong Ji$^{4,5}$, Younes Samih$^6$, Bo Zeng$^1$, Fan Jiang$^1$, Yuanbin Cao$^1$, Dilda Duisenbek$^2$,} \\
    \textbf{Adrian Neo Sau Xun$^2$, Daria Pozdniakova$^2$, Liubou Misevich$^2$, Nevena Marinković$^2$,} \\
    \textbf{Ngoc Gia Linh Nguyen$^2$, Thi Khanh Linh Do$^2$, Sarakmatak Sophy$^2$, Baotian Hu$^8$,} \\
    \textbf{Guanhua Chen$^9$, Gongbo Tang$^2$, Alham Fikri Aji$^7$, Longyue Wang$^1$, Weihua Luo$^1$} \\
    $^1$Alibaba Group
    $^2$Beijing Language and Culture University
    $^3$LMU Munich \\
    $^4$ELLIS Institute Finland
    $^5$University of Turku
    $^6$IBM Research AI, UAE
    $^7$MBZUAI \\
    $^8$Harbin Institute of Technology, Shenzhen
    $^9$Southern University of Science and Technology \\
}

\begin{document}
\maketitle
\begin{abstract}
Large language models (LLMs) are now deployed worldwide, inspiring a surge of benchmarks that measure their multilingual and multicultural abilities.
However, these benchmarks prioritize generic language understanding or superficial cultural trivia, leaving the evaluation of grounded tasks—where models must reason within real-world, context-rich scenarios—largely unaddressed.
To fill this gap, we present \BenchName, a comprehensive and challenging benchmark to assess LLMs’ multilingual and multicultural competence on grounded tasks.
\BenchName is built via a human–AI collaborative framework: expert annotators ensure appropriate difficulty and factual accuracy, while LLMs lighten the manual workload. By incorporating diverse sources, \BenchName ensures comprehensive scenario coverage. Each item is carefully designed to present a high level of difficulty, making \BenchName challenging.
\BenchName contains \SampleNum samples in \LangNum languages of \RegionNum regions, distributed across \TopicNum topics to capture the full breadth of grounded tasks.
The experiments show that the best LLM achieve 44.48\% accuracy on \BenchName,
underscoring substantial room for improvement.\footnote{Code and data are publicly available at \url{https://github.com/AIDC-AI/Marco-LLM}.}

\end{abstract}

\section{Introduction}

\begin{figure}
  \centering
  \begin{subfigure}[t]{\linewidth}   %
    \centering
    \resizebox {\columnwidth} {!} {
    \includegraphics[trim={9cm 12.25cm 11cm 1cm},clip]{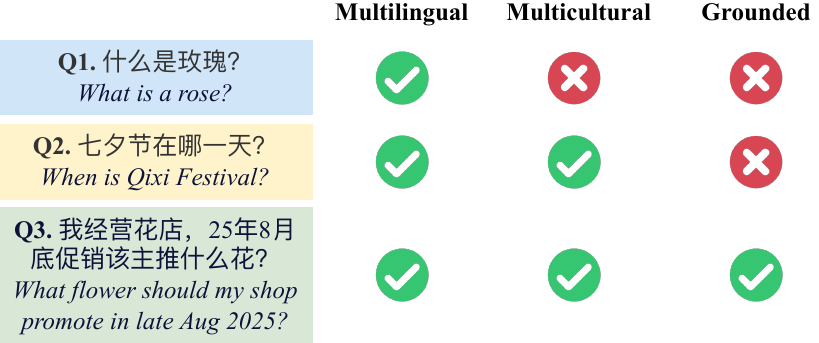}
  }
    \caption{Example-level comparison.}
    \label{fig:example_level}
  \end{subfigure}

  \vspace{3pt}                       %

  \begin{subfigure}[t]{\linewidth}
    \centering
    \resizebox {\columnwidth} {!} {
    \includegraphics[trim={9cm 4.4cm 11cm 10cm},clip]{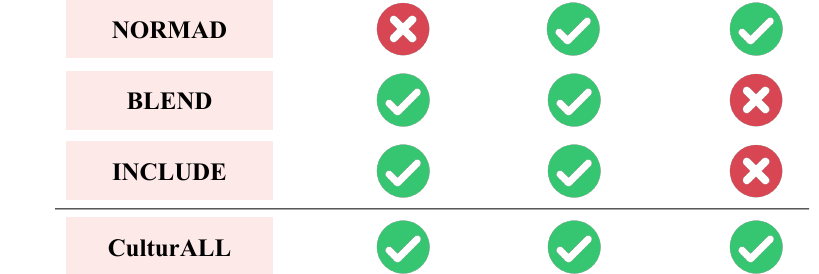}
  }
    \caption{Benchmark-level comparison.}
    \label{fig:benchmark_level}
  \end{subfigure}

  \caption{
(a) Example-level: Q1 is multilingual only; Q2 adds cultural knowledge; Q3 requires all three, posing the hardest challenge.
(b) Benchmark-level: existing representative benchmarks test at most two axes, while \BenchName spans all three.}
  \label{fig:overview}
\end{figure}

As LLMs
are adopted across the globe, it is imperative to evaluate how well they perform in diverse languages and cultures.
Existing multilingual and multicultural benchmarks, e.g., BLEND \citep{DBLP:journals/corr/abs-2406-09948}, INCLUDE \citep{DBLP:journals/corr/abs-2411-19799}, and Global MMLU \citep{DBLP:conf/acl/SinghRFANVLMLSN25}, cover a wide range of languages and cultures, but their content is dominated by encyclopedic trivia. Consequently, they say little about how LLMs perform on the everyday tasks people actually care about, e.g., planning a trip or making an online purchase.
Recent efforts have started to introduce grounded evaluations, e.g., CultureBank \citep{DBLP:conf/emnlp/ShiLZZYHPY24} and NORMAD \citep{DBLP:conf/naacl/RaoYSRS25}.
However, they are mostly in English and cover only a narrow band of grounded tasks, mainly social interactions. This gap prompts a key question: \textit{How effectively can LLMs tackle the diverse grounded tasks users face across different languages and cultures?}

\begin{figure*}
  \centering
  \resizebox {\textwidth} {!} {
    \includegraphics[trim={4.25cm 5cm 4.25cm 4.25cm},clip]{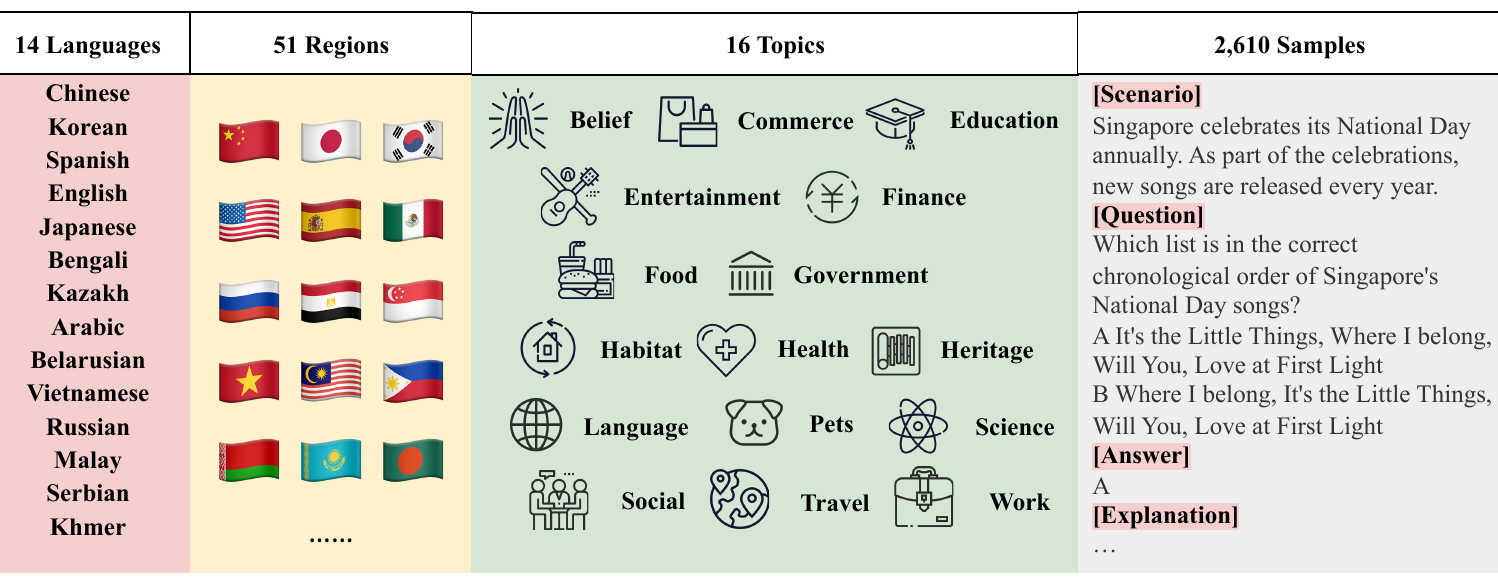}
  }
  \caption{\BenchName is a comprehensive and challenging benchmark. It contains \SampleNum samples in \LangNum languages across \RegionNum regions, distributed among \TopicNum topics to capture the full breadth of grounded tasks. As the given example illustrates, each item presents a grounded scenario followed by its question. Successfully solving each item requires an LLM to fuse these cues with its stored knowledge and reason to the correct answer.}
  \label{fig:multi-culture-bench}
\end{figure*}

A truly capable LLM must solve grounded tasks across diverse linguistic and cultural contexts, because these tasks reflect what users actually need. These tasks are particularly challenging because they probe three complementary capacities of an LLM: \textbf{(1) language comprehension (multilingual)}: the capacity to accurately parse and interpret a user’s native tongue; \textbf{(2) cultural knowledge acquisition (multicultural)}: the ability to access and recall long-tail, domain-specific cultural facts; and \textbf{(3) contextual reasoning (grounded)}: the skill of integrating that information and synthesizing it into an accurate response.
As illustrated in Fig.~\ref{fig:example_level}, Q1 merely tests a LLM's multilingual ability, and Q2 adds a cultural fact. In contrast, Q3 requires the full chain of multilingual, multicultural, and grounded reasoning: the LLM must first interpret the Chinese query, identify the relevant late-August festival in China and its customs, recall the symbolic meanings of different flowers, and finally synthesize this information into a concise, culturally appropriate reply via reasoning. Coordinating this chain of culturally grounded reasoning is anything but trivial.

In response, we introduce \BenchName, the first benchmark to assess LLM performance in grounded scenarios across diverse languages and cultures (Fig.~\ref{fig:benchmark_level}).
\BenchName is constructed using a novel human-LLM collaborative framework that leverages expert annotators for factual accuracy and elevated difficulty, while LLMs assist in generating and enriching diverse scenarios, ensuring comprehensive coverage and challenging samples.
Fig.~\ref{fig:multi-culture-bench} shows \BenchName’s extensive language and cultural coverage, as well as its challenging nature. The characteristics of \BenchName are as follows: 1) coverage: the \SampleNum samples in \BenchName span \TopicNum topics that encompass diverse facets of daily life and society, covering cultures from \RegionNum regions across \LangNum languages; 2) challenging: answering each scenario-based question is difficult because it requires LLMs to integrate nuanced cultural knowledge with strong multi-step reasoning skills.

Using \BenchName, we analyze existing LLMs and find that they struggle with culturally grounded tasks, and improving their performance requires effective web search and strong reasoning capabilities. In summary, our contributions are multifold.

\begin{itemize}
    \item We design a unified human-LLM framework, which can be applied to create benchmarks with wide coverage and high difficulty.
    \item We present \BenchName, the first benchmark explicitly designed to assess LLMs’ multilingual and multicultural competence across a wide spectrum of realistic tasks. %
    \item We benchmark state-of-the-art LLMs on \BenchName and deliver an in-depth analysis, highlighting key strengths and failure modes.
\end{itemize}

\section{\BenchName: Construction and Statistics}

\begin{figure*}
  \centering
  \resizebox {\textwidth} {!} {
    \includegraphics[trim={3.25cm 2cm 3.25cm 2cm},clip]{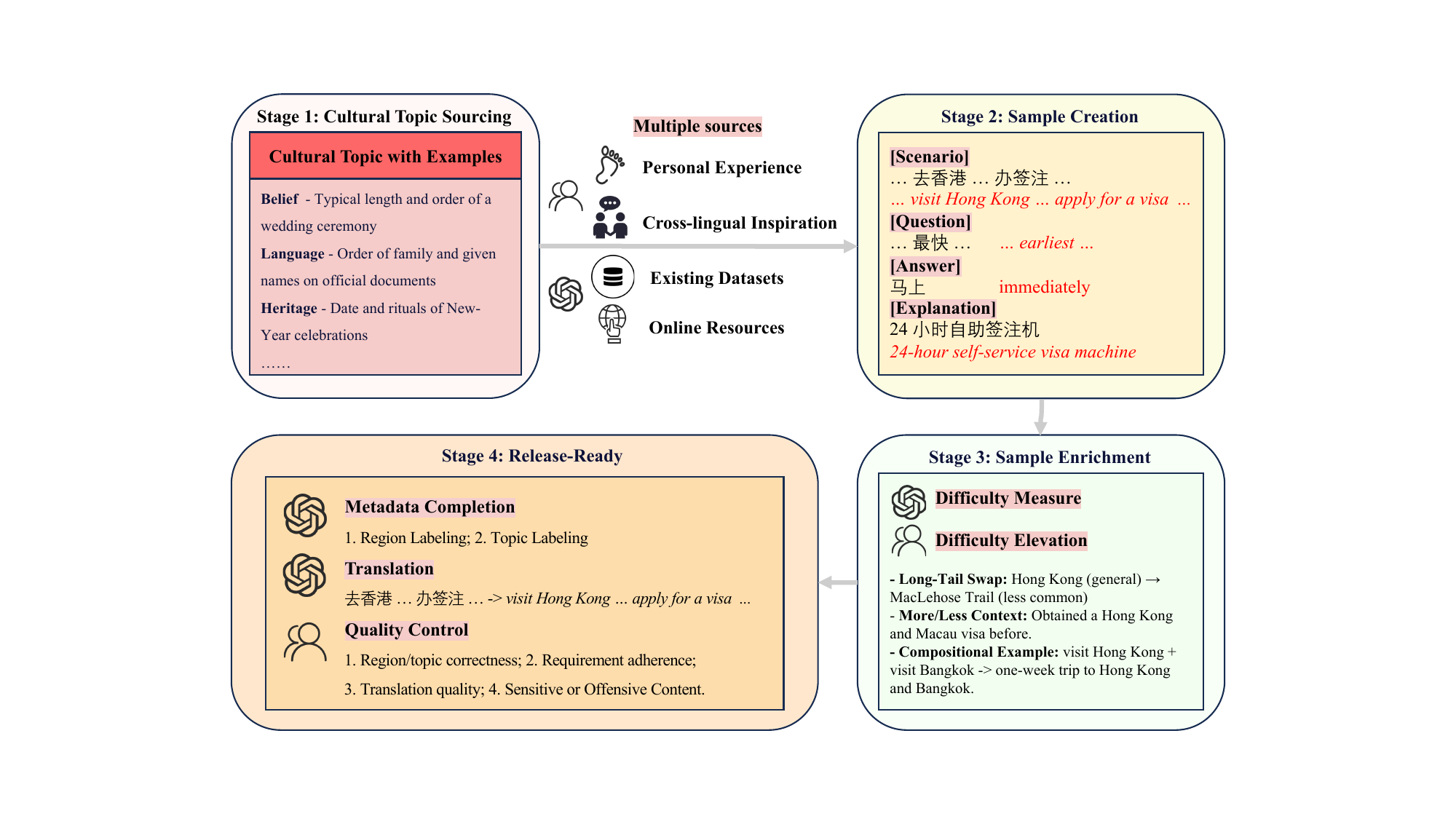}
  }
  \caption{The data construction framework of \BenchName: 1) Cultural Topic Sourcing: assemble a list of cultural topics; 2) Sample Creation: craft original items for each topic; 3) Sample Enrichment: enhance realism and increase difficulty; 4) Release-Ready: complete sample information and conduct quality validation.
  }
  \label{fig:pipeline}
\end{figure*}

Robust evaluation of LLMs on multilingual and multicultural tasks requires datasets that are both diverse and challenging. To achieve this at scale, we introduce a unified human-LLM framework that combines human expertise with the generative power of LLMs, resulting in \BenchName, which offers broad coverage and high difficulty.

An overview of this framework is shown in Fig.~\ref{fig:pipeline}. The framework begins with cultural topic sourcing (\S\ref{sec:task_collection}), compiling an extensive list of cultural topics and illustrative examples. Next is sample creation (\S\ref{sec:sample_creation}), where we draft seed instances for these topics, drawing on sources such as personal experience and online materials. These drafts are then refined during sample enrichment (\S\ref{sec:sample_enrichment}) to increase their difficulty and better mirror grounded scenarios. The final stage—Release-Ready (\S\ref{sec:cross_lingual_augmentation})—completes each sample with topic/region labels and English translation, and then conducts thorough quality checks.

We define a culture group as the population of a single country or region. To capture broad cultural expertise, we collaborate with annotators from a wide range of countries, regions, and linguistic backgrounds. Real-world queries seldom state their cultural origin explicitly, so LLMs must infer it from implicit cues—vocabulary, idioms, institutions, and other context signals. For this reason, each annotator composes samples in the dominant language of their locale—e.g., English in the United States and Mandarin Chinese in mainland China—embedding authentic local references that models must recognize and interpret. All annotation information, including annotator information and guidelines, is provided in \S\ref{sec:annotation}.

\subsection{Stage 1: Cultural Topic Sourcing}
\label{sec:task_collection}

To spur the creation of grounded tasks across a broad spectrum of cultural topics, the cultural-topic-sourcing stage aims to generate a comprehensive list that covers nearly every facet of daily life through human–LLM collaboration. We first compile a preliminary set of topics with concise scope descriptions, drawing on prior research \citep{DBLP:journals/corr/abs-2205-12247,DBLP:journals/corr/abs-2411-19799,DBLP:conf/acl/ChiuJLPLRBATS025} and heuristics, and then engage
\textit{gpt-4o-2024-11-20}
in several iterative rounds to merge, refine, and expand both the topics and their accompanying descriptions.
With the final list in place, we craft seed examples from personal experience and instruct \textit{gpt-4o-2024-11-20} to expand them, yielding a pool of 160 illustrative instances (10 per topic) that serve as scaffolding for the subsequent sample-creation stage. The complete topic list accompanied by descriptions and three representative culture-related scenarios appears in Tab.~\ref{tab:cultural_topics} (\S\ref{sec:topic}); the complete set of examples will be released publicly.

\subsection{Stage 2: Sample Creation}
\label{sec:sample_creation}

\subsubsection{Sample Format}
\label{sec:sample_format}

Tab.~\ref{tab:metadata_fields} outlines the schema that each sample must adhere to. During annotation, the \texttt{language} field is predefined based on the source data or annotator's information, while \texttt{region} and \texttt{topic} are automatically generated by the LLM (see \S\ref{sec:metadata_completion}). All remaining fields are reviewed and completed by the annotators. Below, we detail the requirements for the overall sample and each field that must be verified or completed by annotators.

\begin{table}[t]
\centering
\resizebox {\columnwidth} {!} {
\begin{tabular}{l|l}
\hline
\textbf{Field} & \textbf{Description} \\ \hline
\texttt{language}         & ISO-639-1 code, e.g.\ \texttt{en}, \texttt{zh} \\ \hline
\texttt{region}           & ISO-3166-1 alpha-2 code, e.g.\ \texttt{US}, \texttt{CN} \\ \hline
\texttt{topic}            & One of the \TopicNum\ predefined topics \\ \hline
\texttt{scenario}         & Narrative context in which the question is asked \\ \hline
\texttt{question}         & The question itself \\ \hline
\texttt{answer}           & Correct answer or answer key \\ \hline
\texttt{explanation}      & Concise justification of the answer \\ \hline
\end{tabular}
}
\caption{Metadata for each item.
}
\label{tab:metadata_fields}
\end{table}

\paragraph{Sample} Craft a culturally grounded, grounded item that evaluates an LLM's ability to employ cultural knowledge.
Cultural knowledge includes but not limited to local vocabulary, social norms, cultural commonsense, regulations, and domain-specific knowledge. Generic trivia (e.g., math puzzles or textbook facts) is out of scope.
Two items are considered distinct only if they probe different knowledge or reasoning steps, not if they are merely paraphrases of each other.

\paragraph{Scenario}
Construct a grounded scenario, withholding any explicit hints that would let a model solve the task without relevant cultural knowledge.

\paragraph{Question}
Ensure the query arises from the scenario and cannot be answered correctly without an understanding of the relevant cultural knowledge.

\paragraph{Answer}
To facilitate automatic evaluation, answers should be objective and as brief as possible. If an objective free-form answer is impractical, convert the question to a four-option multiple-choice format (A–D) and return only the chosen letter.

\paragraph{Explanation}
When appropriate, supply the cultural or domain knowledge that supports the answer. These explanations make \BenchName\ more transparent for readers and pave the way for using \BenchName\ in future free-text evaluation tasks.

\subsubsection{Cultural Knowledge Sourcing}
\label{sec:cultural_knowledge_sourcing}

\paragraph{Personal Experience (Human)}
To capture unwritten social cues, emerging slang, and region-specific practices, we ask annotators to draw from their personal experiences. These first-hand contributions result in scenarios that are both authentic and deeply rooted in context. Annotators receive a detailed list of topics with descriptions and examples (\S\ref{sec:task_collection}), which they can adapt, use as inspiration for new culturally relevant instances, or supplement with ideas from local forums.

\paragraph{Cross-lingual Inspiration (Human)}
An example in one language often sparks analogous ideas in annotators who speak other languages. For instance, a Chinese query about obtaining a visa for Hong Kong may inspire a French annotator to create a comparable scenario involving a French employee applying for a Belgian work permit. To facilitate this transfer, we translate existing samples into English (see \S\ref{sec:translation} for details), serving as a shared pivot that enables native speakers of other languages to more easily create parallel data.

\paragraph{Existing Datasets (LLM)}
Many prior cultural benchmarks contain culture-relevant items yet lack explicit grounding in grounded contexts. We refine these items through rewriting using \textit{gpt-4o-2024-11-20}, anchoring each item in a concrete scenario while preserving its original knowledge requirements. Details are provided in \S\ref{sec:existing_datasets}.

\paragraph{Online Resources (LLM)}
We collect culture-rich materials from online resources, focusing primarily on mining posts from Xiaohongshu, guided by our cultural topic example list. For each target country/region (Tab.~\ref{tab:lcode_to_ccode}, \S\ref{sec:existing_datasets}), we combine the region name with topic seeds generated during the cultural-topic–sourcing stage (in Chinese) as search queries to efficiently surface relevant local content. This crawl returns 3,518 pages. Each page is translated into the country/region's dominant language with \texttt{gpt-4o-2024-11-20} (Fig.~\ref{fig:prompt_mt}, \S\ref{sec:prompts}). The retrieved content is then supplied to \textit{gpt-4o-2024-11-20} as raw input for drafting candidate items (Fig.~\ref{fig:prompt_grounded_online}, \S\ref{sec:prompts}).

\subsection{Stage 3: Sample Enrichment}
\label{sec:sample_enrichment}

To ensure \BenchName better reflects challenging grounded demands, every draft item undergoes a process of ``up-leveling'' as illustrated in Fig.~\ref{fig:pipeline}. 
Specifically, we assess the difficulty of the original samples and categorize them into hard and easy examples. Hard examples are forwarded to Stage 4, while easy examples undergo a difficulty elevation process to increase its difficulty if possible.

\subsubsection{Difficulty Measure (LLM)}
\label{sec:difficulty_measure}

Inspired by \citet{phan2025humanity,fabbri2025multinrc}, we utilize three LLMs—\textit{gpt-4o-2024-11-20}, \textit{claude-3.5-sonnet-1022}, and \textit{qwen-max-2024-09-19}—to quantify the difficulty of items produced by Stage 2. An item is classified as challenging if at most one of the three LLMs provides the correct answer. Such items are forwarded to Stage 4. Otherwise, they proceed to the difficulty elevation process for further complexity enhancement.

\subsubsection{Difficulty Elevation (Human)}
\label{sec:difficulty_elevation}
We introduce three complementary enrichment strategies to guide human annotators in elevating the difficulty of existing samples:

\paragraph{Long-Tail Swap}
Common entities are replaced with rarer ones, e.g., substituting the general location "Hong Kong" with "MacLehose Trail," a lesser-known hiking route within the region.

\paragraph{More/Less Context}
Additional situational details are introduced, requiring the answer to hinge on conditional, multi-step reasoning (e.g., determining if a traveler has a prior visa). Conversely, unnecessary context that provides hints to LLMs can be removed to increase the challenge.

\paragraph{Compositional Example}
Two independent knowledge points are combined into a single query—for example, merging the entry requirements for both Hong Kong and Bangkok—forcing the model to engage in compositional reasoning.

Annotators are encouraged to apply one or more of these techniques to enhance sample difficulty. If no opportunities for improvement, annotators may leave it unchanged. These refinements amplify the task complexity, moving beyond superficial matching and compelling models to demonstrate deeper understanding and advanced reasoning capabilities.

\subsection{Stage 4: Release-Ready}
\label{sec:cross_lingual_augmentation}

\subsubsection{Metadata Completion (LLM)}
\label{sec:metadata_completion}

The Metadata Completion step utilizes \textit{gpt-4o-2024-11-20} to fill in the \texttt{region} and \texttt{topic} fields. For \texttt{region}, the prompt shown in Fig.~\ref{fig:region_labeling} (\S\ref{sec:prompts}) is used to generate the corresponding ISO 3166-1 alpha-2 code. For the \texttt{topic}, we prompt the LLM to select the most suitable topic from a predefined list (\S\ref{sec:task_collection}) based on the created sample, using the prompts provided in Fig.~\ref{fig:topic_labeling} (\S\ref{sec:prompts}). %

\subsubsection{Translation (LLM)}
\label{sec:translation}
In this sub-step, \textit{gpt-4o-2024-11-20} translates each sample into English, using the translation prompt in Fig.~\ref{fig:prompt_mt} (\S\ref{sec:prompts}). This process provides a unified reference language for readers while also inspiring annotators to develop new cross-regional samples.

\subsubsection{Quality Control (Human)}
\label{sec:quality_verification}
Data quality is maintained via a peer-review process wherein each annotator cross-checks samples created by others/LLMs. During the review, annotators are asked with checking the following aspects:

\paragraph{Region/Topic Correctness} The assigned region should be a valid ISO 3166-1 alpha-2 code and the topic belongs to the predefined list. Both must accurately align with the content of the sample.
\paragraph{Requirement Adherence} Each sample must comply with the requirements outlined in \S\ref{sec:sample_format}. %
\paragraph{Translation Quality} LLM's translation output should be correctly align with the source input.
\paragraph{Sensitive or Offensive Content} The output should not include personally identifiable information or harmful, offensive, or inappropriate content.

Based on these criteria, annotators have three options: \textbf{accept}, \textbf{revise}, or \textbf{reject}. A sample should be marked as \textbf{accept} if it meets all four criteria. If issues are identified, the annotator should attempt to \textbf{revise} the sample to ensure it fully satisfies the requirements. However, if the sample cannot be revised to meet the criteria, it should be marked as \textbf{reject}. Based on these criteria, 75.0\% of the samples were marked as \textbf{accept}, 8.1\% were marked as \textbf{revise}, and 16.9\% were marked as \textbf{reject}. To ensure robustness, we randomly selected 100 samples from the final dataset for cross-checking, and all of them aligned with the criteria.

\begin{figure*}[t]
    \centering
    \includegraphics[width=\textwidth,trim={2cm 3.5cm 2cm 4cm},clip]{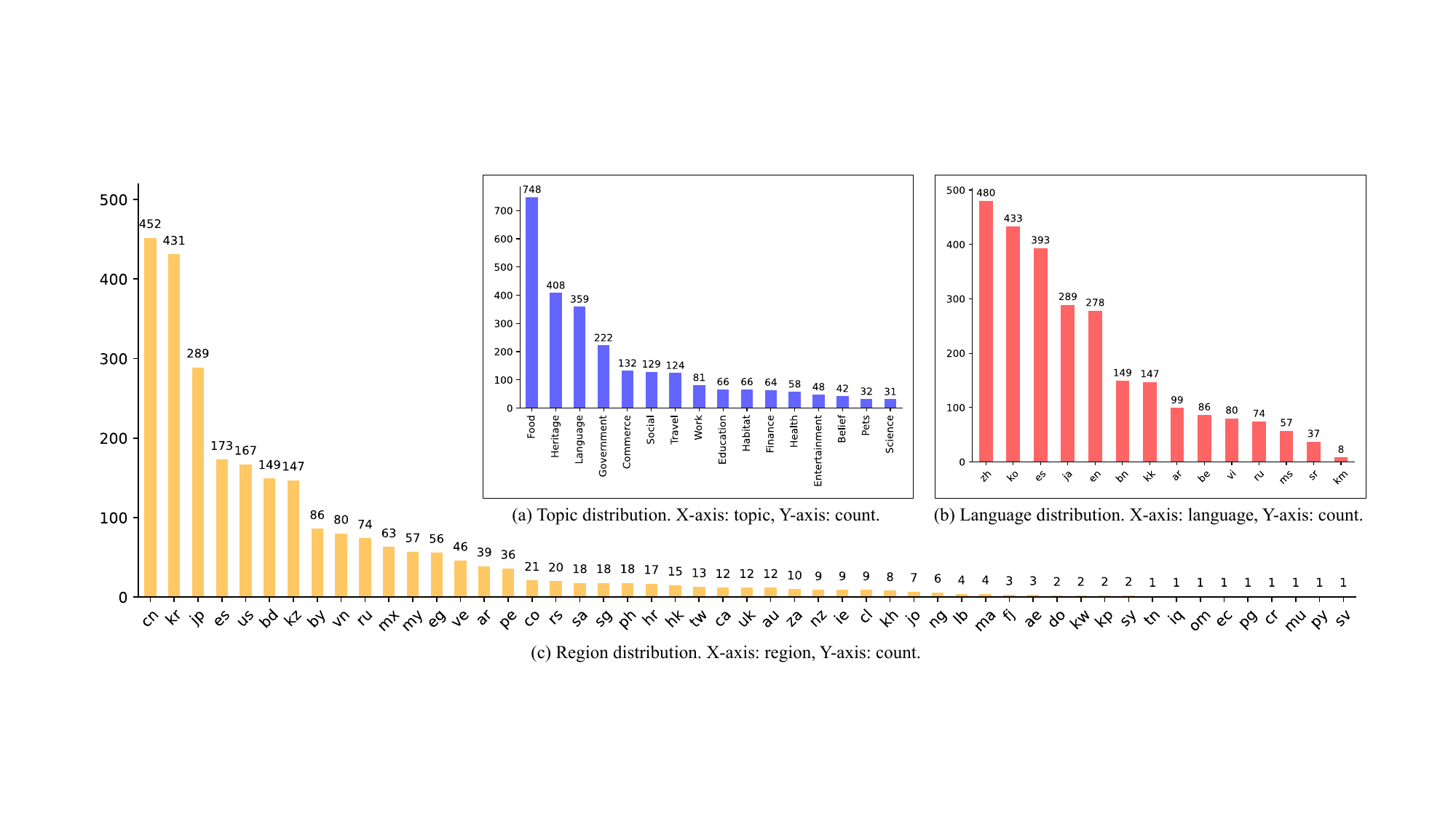}
    \caption{Distributions across topics, languages, and regions. The first row includes: (a) topic distribution and (b) language distribution, and the second row shows (c) region distribution.}
    \label{fig:distributions_two_rows}
\end{figure*}

\subsection{Statistics}
The resulting \BenchName comprises \SampleNum samples, spanning \LangNum languages and \RegionNum regions. The distributions of samples across topics, languages, and regions are illustrated in Fig.~\ref{fig:distributions_two_rows}.

To ensure that \BenchName remains genuinely challenging, it excludes any item that is correctly solved by all 15 model settings (\S\ref{sec:setup}). We categorize the \SampleNum examples in \BenchName based on how many of the 15 settings answered them correctly. Items solved by 10–14 settings are labeled as \textbf{Easy} (470 items, 18.01\%), while those solved by 5–9 settings form the \textbf{Medium} subset (700 items, 26.82\%). Finally, items solved by at most four settings are grouped into the \textbf{Hard} subset (1,440 items, 55.17\%). Fig.~\ref{fig:difficulty} in \S\ref{sec:difficulty_language} further illustrates the language distribution of examples according to the number of settings that answered them correctly.

\section{Setup}
\label{sec:setup}

\paragraph{Model Selection} Tab.~\ref{tab:model_experiment} presents the 15 experiments conducted to benchmark top-performing LLMs on \BenchName. Our analysis includes 8 leading LLMs from the Text Arena leaderboard as of 18 August 2025,\footnote{\url{https://lmarena.ai/leaderboard/text}} spanning both open-source and proprietary models. The experiments feature 15 distinct configurations, achieved by varying reasoning capabilities and the inclusion of web search. This comprehensive experimental setup facilitates a systematic evaluation of key factors, such as reasoning performance, web search integration, model size, and other critical characteristics.

\begin{table*}
\centering
\resizebox{\textwidth}{!} {
\begin{tabular}{l|l|llll|l|l|l|l}
\toprule
\textbf{ID} & \textbf{Experiment Name}                & \textbf{Model Name}        & \textbf{Open}      & \textbf{Reasoning} & \textbf{Web} & \textbf{All} & \textbf{Hard} & \textbf{Med.} & \textbf{Easy} \\
\midrule
1  & gemini-2.5-pro\_auto\_true       & gemini-2.5-pro     & No           & auto            & Yes     & \textbf{44.48}              & \textbf{18.47}              & \textbf{65.71}               & \textbf{92.55}              \\
2  & gemini-2.5-pro\_auto\_false      & gemini-2.5-pro     & No           & auto            & No      & 37.89              & 10.07              & 59.57               & 90.85              \\
3  & gemini-2.5-pro\_128\_true        & gemini-2.5-pro     & No           & 128 tokens      & Yes     & 39.27              & 15.49              & 55.71               & 87.66              \\
4  & gemini-2.5-flash\_auto\_true     & gemini-2.5-flash   & No           & auto            & Yes     & 33.68              & 12.78              & 46.71               & 78.30              \\
5  & gpt-5-20250807\_high\_false      & gpt-5              & No           & high            & No      & 37.59              & 10.28              & 57.71               & 91.28              \\
6  & gpt-5-20250807\_medium\_false    & gpt-5              & No           & medium          & No      & 37.20              & 9.31               & 58.29               & 91.28              \\
7  & gpt-5-20250807\_low\_false       & gpt-5              & No           & low             & No      & 37.24              & 9.38               & 59.00               & 90.21              \\
8  & claude-opus-4-20250514\_high\_false & claude-opus-4      & No           & 1024 tokens     & No      & 36.70              & 9.44               & 56.00               & 91.49              \\
9  & claude-opus-4-20250514\_low\_false  & claude-opus-4      & No           & disabled        & No      & 36.48              & 9.03               & 56.43               & 90.85              \\
10 & claude-sonnet-4-20250514\_high\_false & claude-sonnet-4   & No           & 1024 tokens     & No      & 32.76              & 8.82               & 46.43               & 85.74              \\
11 & qwen-max\_auto\_true             & qwen-max           & No           & hybrid          & Yes     & 19.31              & 5.14               & 24.71               & 54.68              \\
12 & qwen-max\_auto\_false            & qwen-max           & No           & hybrid          & No      & 18.97              & 5.69               & 24.00               & 52.13              \\
13 & qwen3-235b-a22b\_high\_true      & qwen3-235b-a22b    & Yes          & hybrid          & Yes     & 22.49              & 4.24               & 31.43               & 65.11              \\
14 & qwen3-235b-a22b\_high\_false     & qwen3-235b-a22b    & Yes          & hybrid          & No      & 23.68              & 4.65               & 33.57               & 67.23              \\
15 & qwen3-30b-a3b\_high\_true        & qwen3-30b-a3b      & Yes          & hybrid          & Yes     & 17.62              & 4.72               & 23.86               & 47.87              \\
\bottomrule
\end{tabular}
}
\caption{Performance of evaluated LLMs with diverse settings on the complete \BenchName dataset and its three subsets, categorized by difficulty level. All results are reported as accuracy (\%). Reasoning: reasoning capability. Web: the use of web search. Open: open-source availability. Experiment Name: {Model Name}\_{Open}\_{Web}.}
\label{tab:model_experiment}
\end{table*}

\paragraph{Prompt Design} To benchmark LLMs with \BenchName, we conduct zero-shot prompting. In addition, we also require the LLMs to answer the question in as few words as possible to ease the follow-up evaluation. The concrete prompt is provided in Fig.~\ref{fig:run_benchmark} of \S\ref{sec:prompts}.

\paragraph{Metric}
Since all reference answers in \BenchName are strictly objective, automatic assessment can be applied. During evaluation, the judge LLM, \textit{gpt-4o-2024-11-20}, is provided with the full item (scenario, question, gold answer, and optional explanation) alongside the prediction from the evaluated LLM. Leveraging the prompt provided in Fig.~\ref{fig:judge}, the judge assesses whether the prediction aligns with the reference answer. Each item yields a binary outcome (correct or incorrect), and the overall performance is measured as accuracy, defined as the proportion of correctly judged items.

\section{Results}

\subsection{Performance Across LLMs}

The performance of evaluated LLMs across various configurations on the complete \BenchName dataset is detailed in Tab.~\ref{tab:model_experiment}. The resulting scores support several key observations:

\paragraph{Best Setting Still Falls Short}
Among all experiments, gemini-2.5-pro\_auto\_true, which utilizes gemini-2.5-pro with its strongest reasoning capability and web search integration, achieves the highest accuracy
at 44.48\%. However, this performance remains far from ideal, indicating significant room for LLM improvement in handling the challenging scenarios presented by \BenchName.

\paragraph{Open-Source LLMs Lag Behind}
As shown in the results, gemini-2.5-pro (ID 2), gpt-5 (ID 5), and claude-opus-4 (ID 8) achieve accuracy rates of 37.89\%, 37.59\%, and 36.70\%, respectively, demonstrating comparable performance among proprietary models. In contrast, the open-source qwen series shows a significant performance gap, with its top-performing setting, qwen3-235b-a22b\_high\_false (ID 14), achieving only 23.68\%. This disparity highlights the challenges faced by open-source models in addressing multilingual and culturally grounded tasks, particularly when competing with advanced proprietary alternatives.

\paragraph{Higher Variants Consistently Outperform Their Counterparts}
The performance comparisons align with claims about model capabilities. Gemini-2.5-pro (ID 1) outperforms gemini-2.5-flash (ID 4) by 10.80\%, supporting its position as the more powerful variant within the Gemini series. Similarly, claude-opus-4 (ID 8) achieves a 3.94\% higher accuracy than claude-sonnet-4 (ID 10), reinforcing its superior performance. %

\paragraph{Reasoning Effort Affects LLMs Unevenly}
Reasoning capabilities play a crucial role in determining the performance of LLMs, but the impact varies across models. For example, gemini-2.5-pro with the most advanced reasoning setting of "auto" (ID 1) outperforms the same model configured with minimal reasoning (128 tokens, ID 3) by 5.21\%, highlighting the positive effect of enhanced reasoning efforts. However, increasing reasoning capabilities for gpt-5 (ID 5 vs. 6 vs. 7) and claude-opus-4 (ID 8 vs. 9) fails to raise their scores by a considerable margin. This discrepancy could stem from the fact that these LLMs lack sufficiently robust multi-step reasoning abilities to navigate complex cultural scenarios effectively.

\paragraph{Web Search Plays a Critical Role}
Equipping gemini-2.5-pro with a web-search tool raises its score by 6.59\% (ID 1 vs. 2), demonstrating the benefit of external retrieval. In contrast, the qwen models gain little advantage from the web search tool (ID 11 vs. 12, ID 13 vs. 14), suggesting they are not yet sufficiently trained to leverage web-search results for solving grounded tasks.

\subsection{Performance Across Difficulty Levels}

As shown in Tab.~\ref{tab:model_experiment}, the performance of LLMs across the three subsets highlights several key patterns. First, the relative ranking of different experimental settings remains consistent across Easy, Medium, and Hard tasks, indicating stable differences in their capabilities. Second, the performance gap between commercial and open-source models becomes more pronounced as tasks grow easier: on the Easy subset, gemini-2.5-pro\_auto\_true (ID 1) achieves an impressive accuracy of 92.55\%, while the best-performing open-source model (ID 14) achieves only 67.23\%, trailing by a substantial margin of 25.32\%. Finally, all systems continue to struggle with the Hard subset; even the top-performing setting (ID 1) achieves 18.47\%, highlighting the need for further advancements.

\begin{figure*}
    \centering
    \includegraphics[width=\textwidth]{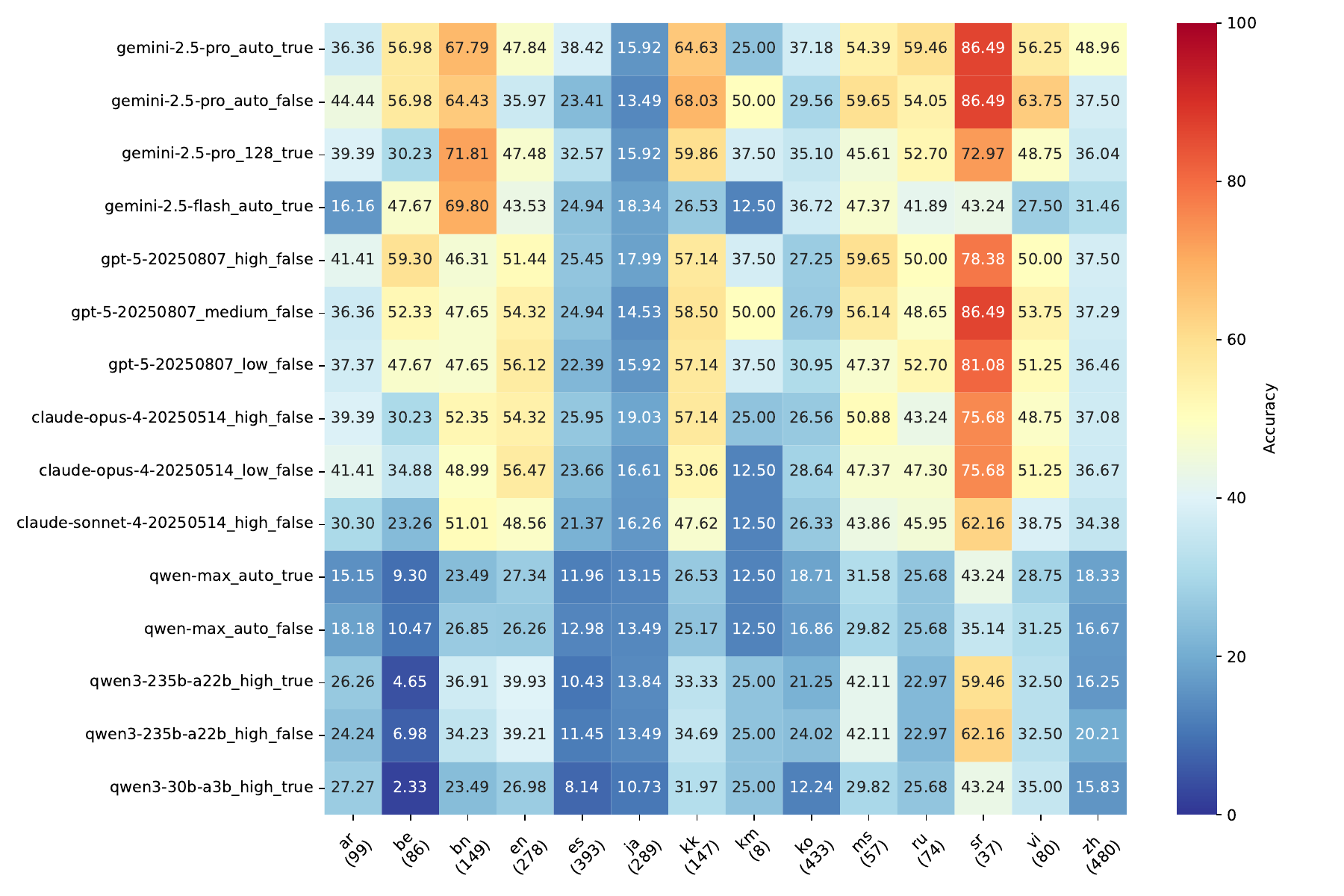}
    \caption{Performance of various experimental settings across \LangNum languages. X-axis: languages (along with their sample counts), Y-axis: different experimental settings.}
    \label{fig:language_performance}
\end{figure*}

\subsection{Performance Across Languages}

We further investigate the performance of different experimental settings across languages, as depicted in the heatmap in Fig.~\ref{fig:language_performance}. Certain languages, such as Serbian, are relatively less challenging, whereas others, like Japanese, pose greater difficulties. Notably, the performance rankings of languages vary significantly across different experimental settings. For example, gemini-2.5-flash\_auto\_true achieves an accuracy of 69.80\% in Bengali, demonstrating competitive results against state-of-the-art settings. However, in other languages, e.g., Arabic and Chinese, its performance lags significantly behind alternative configurations.

\subsection{English vs. In-Language Prompts}

Previous work \citep{DBLP:journals/corr/abs-2205-12247} revealed that samples written in English tend to achieve superior performance compared to those written in native languages, as LLMs typically exhibit greater proficiency in English. In our study, we evaluated the performance using native prompts and their English translations under the experimental setting of ID 1. The results show that employing English prompts yields an accuracy of 36.40\%, which is 8.08\% lower than the accuracy achieved with the original native prompts. We hypothesize that this discrepancy arises because the native language inherently reflects the cultural context of the scenario, whereas translation may dilute or lose these nuances. These findings highlight the urgent need to enhance LLMs' capabilities to adapt to diverse cultural and linguistic contexts.

\section{Related Work}

A wide range of benchmarks have been have been introduced to assess the general capabilities of LLMs \citep{DBLP:conf/iclr/HendrycksBBZMSS21,DBLP:conf/nips/WangMZNCGRAHJLK24}. As LLMs become ubiquitous across the world, researchers are paying growing attention to their performance across diverse languages \citep{DBLP:journals/corr/abs-2504-15521} and cultures \citep{DBLP:conf/acl/HershcovichFLLA22,DBLP:conf/emnlp/AdilazuardaMLSA24,DBLP:journals/corr/abs-2411-00860,DBLP:journals/tacl/LiuGK25}.

Early efforts to assess LLMs under multilingual and multicultural settings have curated knowledge bases and probing tasks to test the LLMs' capacity on culture-specific knowledge acquisition \citep{DBLP:journals/corr/abs-2205-12247,DBLP:conf/naacl/WangLHJDAC24,DBLP:journals/corr/abs-2406-09948,DBLP:conf/naacl/ZhouKLGCCLH25,DBLP:conf/acl/HasanHALUSKCA25,DBLP:conf/acl/AroraKCBIC25}. Findings from these studies show that LLMs still display pronounced cultural biases and uneven performance across different regions of the world \citep{DBLP:conf/acl/NaousRR024,DBLP:conf/naacl/MitchellABCCDDHDDDGFHKL25}. While these benchmarks shed light on what LLMs know about diverse cultures in different languages, they do not fully assess multilingual and multicultural competence, which requires LLMs not only to store cultural knowledge but also to apply it flexibly in grounded scenarios \citep{DBLP:conf/naacl/RaoYSRS25}.

To obtain a clearer picture of LLMs’ multilingual and multicultural competence, recent benchmarks have shifted from decontextualized trivia to grounded scenarios, covering social interaction \citep{DBLP:conf/naacl/RaoYSRS25,DBLP:conf/nips/YinQHC024,DBLP:conf/naacl/QiuFAHTPW25}, psychology \citep{DBLP:conf/iclr/0001KPL0AOSSM0S25}, and cultural proverbs \citep{DBLP:conf/naacl/LiuKBG24}. However, they are usually monolingual or confined to a narrow domain, leaving them a long way from the breadth and diversity of situations that arise in grounded applications.

\section{Conclusion}

In this paper, we introduce \BenchName, the first benchmark designed to evaluate the multilingual and multicultural capabilities of LLMs across grounded tasks. \BenchName is built using a human-LLM collaboration framework, ensuring both comprehensiveness and a high level of challenge. Through an in-depth analysis of LLMs on \BenchName, we highlight the critical need to enhance their information retrieval and reasoning skills.

\section*{Limitations}

The scope and design of this study inevitably come with certain limitations, which we outline in this section. Addressing these limitations in future work can help provide a more nuanced and comprehensive understanding of LLMs' performance across cultural contexts.

\paragraph{Coverage Bias}
As illustrated in Fig.~\ref{fig:distributions_two_rows}, certain topics, languages, and regions are underrepresented in the dataset. Nevertheless, our proposed unified data construction framework provides a practical foundation for expanding \BenchName to improve coverage in the future.

\paragraph{Focus on Regional Cultural Groupings}
In this work, we define cultural groups predominantly based on geographic regions. While this approach provides a high-level understanding of general cultural differences, it does not fully capture more nuanced or cross-cutting aspects of culture. Factors such as religion, age, socio-economic status, gender, and education significantly influence cultural perspectives and may intersect in ways that are not represented by regional groupings alone. Future studies should explore these more fine-grained cultural dimensions to offer a holistic assessment of LLMs' cultural-grounded reasoning capabilities.

\paragraph{Reliance on Objective Answer Evaluation}
To ensure consistency and reproducibility in our experimental setup, we focus exclusively on tasks with objective, verifiable answers to enable automatic evaluation. While these tasks serve as a robust benchmark for model performance, they do not account for the complexities of free-text generation, which is a key feature of LLMs in multilingual and culturally nuanced applications. Investigating free-text generation compared to objective reasoning tasks is an important avenue for future exploration to better understand LLMs' ability to engage with subjective, open-ended questions influenced by cultural relativism.

\paragraph{Exclusion of Multimodal Inputs}
Our study focuses entirely on text-based inputs without considering multimodal contexts, such as the integration of visual, auditory, or other non-textual signals. However, cultural understanding often extends beyond textual communication to include visual symbolism, nonverbal gestures, and audio cues, all of which hold significant meaning in cultural interactions. Future research should explore the impact of multimodality on LLM performance when tackling culturally grounded tasks to better model the complexities of human communication.

\section*{Ethical Considerations}

\paragraph{Annotation Process and Annotator Profile}
Prior to starting the annotation process, annotators undergo a comprehensive briefing on the guidelines detailed in \S\ref{sec:annotation_guideline}. Each annotator must first label a pilot batch of 10 randomly selected examples. Only those who complete this trial accurately—demonstrating full comprehension of the guidelines—may advance to the main annotation phase. During production we run continuous spot-checks and feedback rounds to keep quality high.

The annotation team consists of native speakers or individuals with extensive immersion in the target language (see \S\ref{sec:annotators} for demographics). Annotators were informed about the purpose of the data collection, its intended use, and storage policies through detailed instructions and a privacy agreement.

\paragraph{Reproducibility Challenges and Mitigation Strategies}

To facilitate reproducibility, we will make publicly available: (i) the key code components used for data collection, processing, and evaluation; (ii) the finalized \BenchName\ dataset accompanied by detailed documentation of its construction and evaluation workflow; and (iii) the full set of model outputs, performance scores, and all experimental configurations required to replicate our results.

All artifacts will be made publicly available at the time of publication, enabling anyone to fully reproduce the entire workflow, from raw inputs to the final results presented in our tables.

\bibliography{custom}

\appendix

\section{Annotation}
\label{sec:annotation}

\subsection{Annotators}
\label{sec:annotators}

\begin{table}[t]
\centering\small
\renewcommand{\arraystretch}{1.1}
\begin{tabularx}{\columnwidth}{@{}l c c >{\raggedright\arraybackslash}X@{}}
\toprule
\textbf{ID} & \textbf{Nationality} & \textbf{Target Lang.} & \textbf{Background} \\
\midrule
A & Chinese      & Chinese    & Native speaker \\
B & Chinese      & Chinese    & Native speaker \\
C & Bangladeshi  & Bengali    & Native speaker \\
D & Kazakh       & Kazakh     & Native speaker \\
E & Moroccan     & Arabic     & Native speaker \\
F & Belarusian   & Belarusian & Native speaker \\
G & Vietnamese   & Vietnamese & Native speaker \\
H & Russian      & Russian    & Native speaker \\
I & Malaysian    & Malay      & Native speaker \\
J & Serbian      & Serbian    & Native speaker \\
K & Cambodian    & Khmer      & Native speaker \\
L & Chinese      & Korean     & Ten years’ study \\
M & Chinese      & Spanish    & Eight years’ study \\
N & Chinese      & English    & Twenty years’ study; five years’ residence \\
O & Chinese      & Japanese   & Ten years’ study; one year’s residence \\
\bottomrule
\end{tabularx}
\caption{Annotator profiles: nationality, target language, and background with the target language.}
\label{tab:evaluator-profile}
\end{table}

We relied on 15 volunteer annotators from universities and industrial research labs. All volunteers are native speaker of the target language, or near-native with more than 5 consecutive years of residence and study in the language community. Table~\ref{tab:evaluator-profile} gives an overview of their backgrounds.

\subsection{Guideline}
\label{sec:annotation_guideline}

As illustrated in Fig.~\ref{fig:pipeline}, human annotators are assigned four distinct tasks, which are detailed below.

\subsubsection{Task~A: Sample Creation (Personal Experience)}

\begin{enumerate}
    \item Read \S\ref{sec:sample_format} to fully understand the sample requirements.
    \item Browse the full topic list—including descriptions and seed examples (\S\ref{sec:task_collection})—to select a topic that interests you.
    \item Craft an original sample based on your personal experience whenever possible, using seed items and local forums as inspiration.
\end{enumerate}

\subsubsection{Task~B: Sample Creation (Cross-lingual Inspiration)}

\begin{enumerate}
    \item Read \S\ref{sec:sample_format} to fully understand the sample requirements.
    \item Read the English translations of examples originally created in other languages.
    \item Whenever possible, write a culturally plausible example in your native language that is similar to the provided ones.
\end{enumerate}

\subsubsection{Task~C: Difficulty Elevation}

\begin{enumerate}
    \item Review \S\ref{sec:sample_format} to fully understand the sample requirements and \S\ref{sec:difficulty_elevation} to learn strategies for increasing difficulty.
    \item Retrieve an easy sample for your language and region.
    \item If possible, enhance its difficulty using the elevation techniques described in \S\ref{sec:difficulty_elevation}.
\end{enumerate}

\subsubsection{Task~D: Quality Control}
\label{sec:task4}

\begin{enumerate}
    \item Verify the sample against three criteria: Region/Topic Correctness, Requirement Adherence, Translation Quality, and Sensitive or Offensive Content, as outlined in \S\ref{sec:quality_verification}.
    \item \textbf{Accept} the sample if it fully satisfies all three criteria without any issues.
    \item If issues are identified, \textbf{revise} the sample to ensure it meets all requirements.
    \item If the sample cannot be revised to meet the criteria, it should be marked as \textbf{reject}.
\end{enumerate}

\section{Topics}
\label{sec:topic}

Tab.~\ref{tab:cultural_topics} presents the full topic list, each entry paired with a short description and three illustrative culture-specific scenarios; the complete set of seed examples will be released publicly.

\begin{table*}
\scriptsize
\setlength\tabcolsep{4pt}          %
\renewcommand\arraystretch{1.15}   %
\centering
\begin{tabularx}{\textwidth}{p{0.095\textwidth}|X|%
                             >{\raggedright\arraybackslash}p{0.17\textwidth}|%
                             >{\raggedright\arraybackslash}p{0.17\textwidth}|%
                             >{\raggedright\arraybackslash}p{0.17\textwidth}}
\hline
\textbf{Topic} & \textbf{Description} & \textbf{Example 1} & \textbf{Example 2} & \textbf{Example 3} \\ \hline
Belief &
Systems of conviction that shape values, rituals, institutions, life-cycle events, and views on existence—covering religious faith, spiritual practice, secular ethics, and cultural traditions (e.g., funerary customs and ideas of an afterlife). &
Typical length and order of a wedding ceremony &
Dietary restrictions during major religious holidays &
Whether to pull the lever in the classic trolley-problem dilemma \\ \hline
Commerce &
Buying, selling, marketing, and payment of goods and services—from daily necessities to luxury fashion—across bricks-and-mortar shops, e-commerce sites, and mobile wallets. &
Typical opening hours for supermarkets &
Return policy for online purchases &
Legal limits on alcohol sales in retail stores \\ \hline
Education &
Formal and informal learning, teaching, research, and skill-building for all ages, settings, and disciplines. &
Courses normally taken in middle school &
National university-entrance-exam format &
Grading scale used in secondary schools \\ \hline
Entertainment &
Media, arts, sports, games, performances, hobbies, and events created for leisure and enjoyment. &
Popular sport clubs &
National mascots or iconic cartoon characters &
Gambling age and casino legality \\ \hline
Finance &
Earning, saving, budgeting, investing, insuring, transferring, and distributing wealth during life and after death. &
Color that signals a stock-price rise or fall on trading screens &
Common payment methods in everyday shopping &
Typical tax-filing deadline for individuals \\ \hline
Food &
Agriculture, sourcing, processing, cooking, nutrition, beverages, and dining culture from farm to table. &
Typical breakfast foods &
Is tipping expected in restaurants? &
Common allergens that must be listed on packaged food \\ \hline
Government &
Public policy, legislation, courts, law enforcement, defense, emergency response, and civic administration. &
Highway speed limits &
Emergency number to call when lost in the mountains &
Length of mandatory military or civil service \\ \hline
Habitat &
Homes, buildings, infrastructure, utilities, urban planning, ecosystems, weather patterns, and sustainability practices. &
Typical home-heating system &
Floor-numbering convention in multi-story buildings &
Recycling rules for household waste \\ \hline
Health &
Physical, mental, and emotional well-being—prevention, treatment, fitness, wellness, palliative, and end-of-life care. &
Standard childhood-vaccination schedule &
Prescription vs.\ over-the-counter drug availability &
Legal age of consent for medical decisions \\ \hline
Heritage &
Past events, living traditions, festivals, monuments, and other cultural inheritances—and their study, preservation, and commemoration. &
Date and rituals of New-Year celebrations &
Historic event marked by a public holiday &
Customs from a particular historical period \\ \hline
Language &
Official and minority languages, scripts, dialects, idioms, emotional nuance, politeness levels, sign language, literacy, and translation norms. &
Order of family and given names on official documents &
Appropriate greetings and honorifics in business &
Meaning and proper use of a common proverb \\ \hline
Pets &
Care, health, training, companionship, and welfare of domesticated animals. &
Rules for bringing pets on public transport &
Mandatory rabies vaccination for dogs &
Cultural status of certain animals \\ \hline
Science &
Systematic inquiry into the natural world and its applications—research, engineering, technology, and innovation. &
Unit used to state distance between two cities &
Standard format for writing dates &
Whether smartphones support dual-SIM use \\ \hline
Social &
Family, friendships, romance, community networks, demographics, and social issues. &
Table etiquette at family gatherings &
Meaning of two women holding hands in public &
Typical blind-dating process \\ \hline
Travel &
Planning, transport, logistics, accommodation, tourism, and movement of people or goods. &
Information needed before booking a city trip &
Visa rules for a 90-day tourist stay &
Cost of popular tourist attractions \\ \hline
Work &
Careers, labor markets, workplaces, productivity tools, and professional development. &
Statutory length of paid annual leave &
Legal steps for ending an employment contract &
Region-specific unique occupations \\ \hline
\end{tabularx}
\caption{Cultural topics with concise descriptions and illustrative examples.}
\label{tab:cultural_topics}
\end{table*}

\section{Adapting Existing Datasets}
\label{sec:existing_datasets}

\paragraph{Data Sources}
We repurpose six public benchmarks:

\begin{itemize}
\item \textbf{GEOMLAMA} \citep{DBLP:journals/corr/abs-2205-12247}: QA pairs in five language–region pairs.
\item \textbf{SeaEval} \citep{DBLP:conf/naacl/WangLHJDAC24}: multiple-choice questions in four language–region pairs.
\item \textbf{INCLUDE} \citep{DBLP:journals/corr/abs-2411-19799}: 44 languages.
We retain items whose \texttt{regional\_feature} is \textit{region implicit}, \textit{region explicit}, or \textit{culture}.
\item \textbf{CulturalBench} \citep{DBLP:conf/acl/ChiuJLPLRBATS025}: 45 countries/regions, English only. Each item is translated into the dominant local language (Tab.~\ref{tab:ccode_to_lcode}).
\item \textbf{Global-MMLU} \citep{DBLP:conf/acl/SinghRFANVLMLSN25}: 42 languages. We 
keep only culture-sensitive questions.
\item \textbf{MultiNRC} \citep{DBLP:journals/corr/abs-2507-17476}: QA pairs in three language–region pairs.
\end{itemize}

\paragraph{Translation}
Whenever an item is not already in the target language, we translate it with \texttt{gpt-4o-2024-11-20} using the prompt shown in Fig.~\ref{fig:prompt_mt} (§\ref{sec:prompts}).

\paragraph{Grounding}
The translated (or original) text is then converted into our sample format with the prompt in Fig.~\ref{fig:prompt_grounded_existing}, ensuring that each item remains original cultural knowledge while satisfying \BenchName's annotation schema.

\begin{table}[t]
\centering
\begin{tabular}{llll}
\toprule
\textbf{Lang} & \textbf{Language} & \textbf{Reg} & \textbf{Region} \\
\midrule
ar & Arabic       & sa & Saudi Arabia       \\
be & Belarusian   & by & Belarus            \\
bn & Bengali      & bd & Bangladesh         \\
en & English      & us & United States      \\
es & Spanish      & es & Spain              \\
ja & Japanese     & jp & Japan              \\
kk & Kazakh       & kz & Kazakhstan         \\
km & Khmer        & kh & Cambodia           \\
ko & Korean       & kr & South Korea        \\
ms & Malay        & my & Malaysia           \\
ru & Russian      & ru & Russia             \\
sr & Serbian      & rs & Serbia             \\
vi & Vietnamese   & vn & Vietnam            \\
zh & Chinese      & cn & China              \\
\bottomrule
\end{tabular}
\caption{ISO 639-1 language codes (Lang) with language names and their representative ISO 3166-1 alpha-2 country/region codes (Reg) and region names, sorted by Lang.}
\label{tab:lcode_to_ccode}
\end{table}

\begin{table*}[t]
\centering
\scriptsize
\setlength{\tabcolsep}{2pt}
\resizebox{\textwidth}{!}{
\begin{tabular}{llll|llll|llll}
\toprule
\textbf{Reg} & \textbf{Region} & \textbf{Lang} & \textbf{Language} &
\textbf{Reg} & \textbf{Region} & \textbf{Lang} & \textbf{Language} &
\textbf{Reg} & \textbf{Region} & \textbf{Lang} & \textbf{Language} \\
\midrule
ar & Argentina        & es & Spanish     & au & Australia        & en & English      & bd & Bangladesh      & bn & Bengali      \\
by & Belarus          & be & Belarusian  & ca & Canada           & en & English      & cl & Chile           & es & Spanish      \\
cn & China            & zh & Chinese     & co & Colombia         & es & Spanish      & eg & Egypt           & ar & Arabic       \\
es & Spain            & es & Spanish     & fj & Fiji             & en & English      & hr & Croatia         & hr & Croatian     \\
hk & Hong Kong        & zh & Chinese     & ie & Ireland          & en & English      & iq & Iraq            & ar & Arabic       \\
jp & Japan            & ja & Japanese    & kh & Cambodia         & km & Khmer        & kp & North Korea     & ko & Korean       \\
kr & South Korea      & ko & Korean      & kz & Kazakhstan       & kk & Kazakh       & lb & Lebanon         & ar & Arabic       \\
ma & Morocco          & ar & Arabic      & mu & Mauritius        & en & English      & mx & Mexico          & es & Spanish      \\
my & Malaysia         & ms & Malay       & ng & Nigeria          & en & English      & nz & New Zealand     & en & English      \\
pe & Peru             & es & Spanish     & pg & Papua N. Guinea  & en & English      & ph & Philippines     & tl & Tagalog      \\
rs & Serbia           & sr & Serbian     & ru & Russia           & ru & Russian      & sa & Saudi Arabia    & ar & Arabic       \\
sg & Singapore        & en & English     & sy & Syria            & ar & Arabic       & tw & Taiwan          & zh & Chinese      \\
uk & United Kingdom   & en & English     & us & United States    & en & English      & ve & Venezuela       & es & Spanish      \\
vn & Vietnam          & vi & Vietnamese  & za & South Africa     & en & English      &    &                 &    &              \\
\bottomrule
\end{tabular}
}
\caption{ISO~3166-1 alpha-2 country/region codes (Reg) and region names with their representative ISO~639-1 language codes (Lang) and language names, sorted alphabetically by Reg.}
\label{tab:ccode_to_lcode}
\end{table*}

\section{Prompts}
\label{sec:prompts}

This section lists the prompts employed in our study, including translation (Fig.~\ref{fig:prompt_mt}), grounded sample generation based on existing datasets (Fig.~\ref{fig:prompt_grounded_existing}), grounded sample generation based on online resources (Fig.~\ref{fig:prompt_grounded_online}), region labeling (Fig.~\ref{fig:region_labeling}), topic labeling (Fig.~\ref{fig:topic_labeling}), \BenchName evaluation (Fig.~\ref{fig:run_benchmark}), and prediction judgment (Fig.~\ref{fig:judge}).

\begin{figure}
  \centering
  \begin{tcolorbox}[title={Translation}]
    Translate the following text to \texttt{\{target\_language\}}:

    \texttt{\{source\}}

    Do not output anything else.
  \end{tcolorbox}
  \caption{Prompt used for the translation task, where \texttt{\{target\_language\}} is the target language name and \texttt{\{source\}} is the input source text for translation.}
  \label{fig:prompt_mt}
\end{figure}

\begin{figure}
  \centering
  \begin{tcolorbox}[title={Grounded Sample Generation (Existing Datasets)}]
You are given a sample describing some cultural knowledge:

\texttt{\{source\_excerpt\}}

Generate a grounded item (scenario + question + answer + explanation) that assesses the consultant’s grasp of the cultural knowledge. 

Ensure the generated item preserves the same cultural knowledge as the example. Do not modify the choices or the correct answer.

The generated sample should be in the same language as the given sample.

The output format should be:

[Scenario]

XXX  

[Question]  

XXX  

[Answer]  

XXX  

[Explanation]  

XXX  

Do not output any other things.
  \end{tcolorbox}
  \caption{Prompt used for grounded‐sample creation. Placeholders \texttt{\{source\_topic\}}, \texttt{\{source\_excerpt\}}, and \texttt{\{topic\_list\}} are replaced with the corresponding inputs.}
  \label{fig:prompt_grounded_existing}
\end{figure}

\begin{figure}
  \centering
  \begin{tcolorbox}[title={Grounded Sample Generation (Online Resources)}]
You are given a sample describing some cultural knowledge:

\texttt{\{source\_excerpt\}}

Generate a grounded item (scenario + question + answer + explanation) that assess the consultant's grasp of the cultural knowledge.  

Ensure the generated sample preserves the same cultural knowledge as the provided example.

The answer should be objective and as brief as possible. If an objective free-form answer is impractical, convert the question to a four-option multiple-choice format (A–D) and return only the chosen letter.

The generated sample should be in the same language as the given sample.

The output format should be:

[Scenario]

XXX  

[Question]

XXX  

[Answer] 

XXX  

[Explanation] 

XXX  

Do not output any other things.
  \end{tcolorbox}
  \caption{Prompt used for grounded-sample creation from online resources. The placeholder \texttt{\{source\_excerpt\}} is replaced with the source text.}
  \label{fig:prompt_grounded_online}
\end{figure}

\begin{figure}
  \centering
  \begin{tcolorbox}[title={Region Labeling}]
    Based on the following text and its written language,
    retrieve the corresponding ISO 3166-1 alpha-2 country code (only one lowercase two-letter country code).
    
    Scenario: \texttt{\{scenario\}}
    
    Question: \texttt{\{question\}}
    
    Do not output any other things.
  \end{tcolorbox}
  \caption{Prompt used for region classification, where \texttt{\{scenario\}} and \texttt{\{question\}} are the provided fields of the given sample.}
  \label{fig:region_labeling}
\end{figure}

\begin{figure}
  \centering
  \begin{tcolorbox}[title={Topic Labeling}]
    Based on the following text,
    select the most appropriate topic from the topic list \texttt{\{topic\_list\}}.
    
    Scenario: \texttt{\{scenario\}}
    
    Question: \texttt{\{question\}}
    
    Do not output any other things.
  \end{tcolorbox}
  \caption{Prompt used for topic classification. \texttt{\{scenario\}} and \texttt{\{question\}} are the provided fields of the given sample. \texttt{\{topic\_list\}} is the predefined list.}
  \label{fig:topic_labeling}
\end{figure}

\begin{figure}
  \centering
  \begin{tcolorbox}[title={\BenchName Evaluation}]
    Using the scenario as context, answer the question in as few words as possible.
    
    Scenario: \texttt{\{scenario\}}
    
    Question: \texttt{\{question\}}
    
    Answer:
  \end{tcolorbox}
  \caption{Prompt used for \BenchName evaluation, where \texttt{\{scenario\}} and \texttt{\{question\}} are the provided fields of the given sample.}
  \label{fig:run_benchmark}
\end{figure}

\begin{figure}
  \centering
  \begin{tcolorbox}[title={Judgement}]
    Please evaluate whether the model prediction is correct based on the given scenario, question, answer, and explanation.

    \textbf{Scenario:} \texttt{\{scenario\}} \\
    \textbf{Question:} \texttt{\{question\}} \\
    \textbf{Answer:} \texttt{\{answer\}} \\
    \textbf{Explanation:} \texttt{\{explanation\}} \\

    \textbf{Model prediction:} \texttt{\{prediction\}}

    Only output \texttt{1} or \texttt{0} with no additional text. \texttt{1} means correct, \texttt{0} means incorrect.
  \end{tcolorbox}
  \caption{Prompt used to evaluate the model's prediction based on the given scenario, question, answer, and explanation. The prompt ensures binary evaluation for correctness.}
  \label{fig:judge}
\end{figure}

\section{Difficulty Distribution Across Languages}
\label{sec:difficulty_language}

\begin{figure}
    \centering
    \includegraphics[width=\columnwidth]{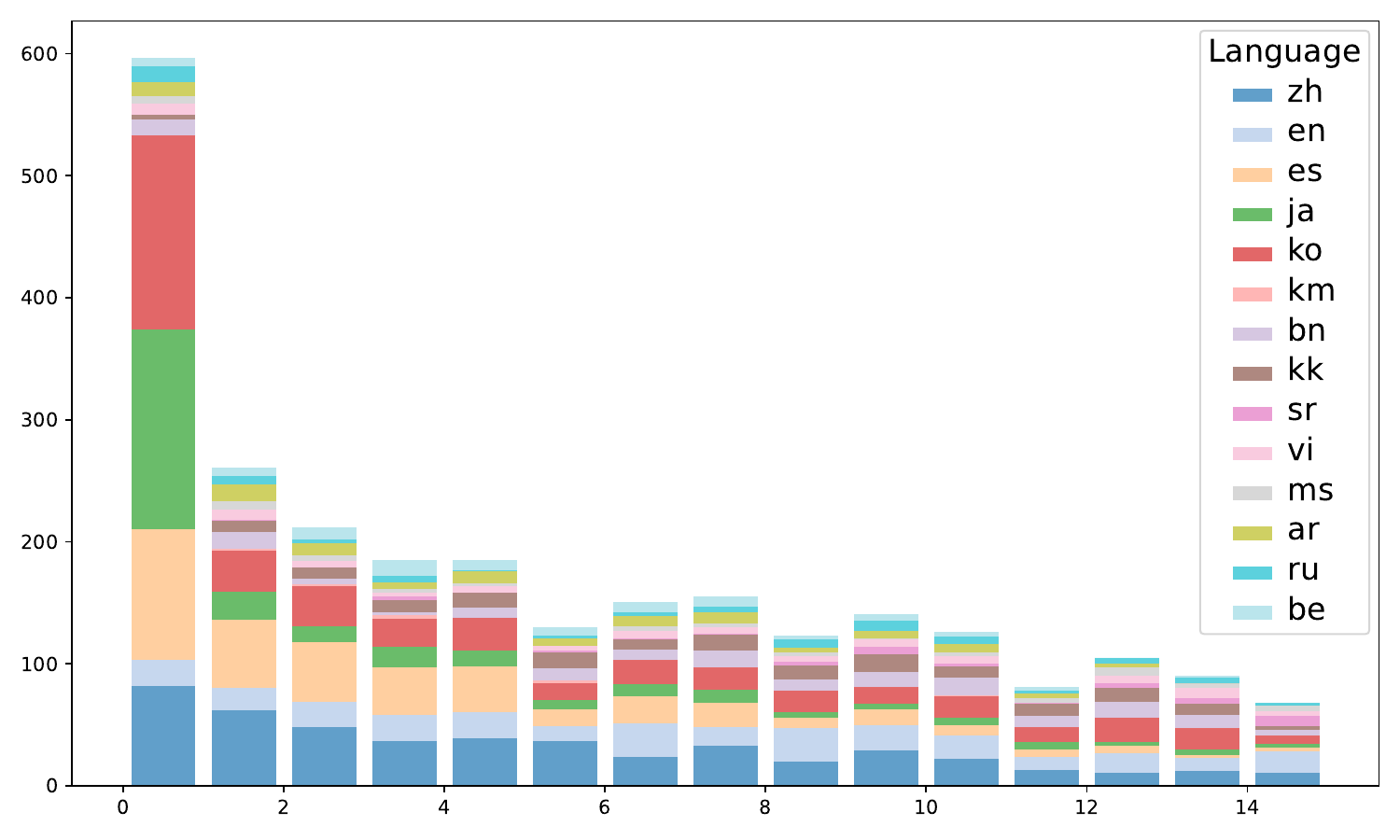}
    \caption{Language distribution of examples based on the number of settings that answered them correctly. X-axis: Number of settings that answered correctly, Y-axis: Count of examples.}
    \label{fig:difficulty}
\end{figure}

\end{document}